\documentclass[10pt, a4paper]{article}
\usepackage{lrec2006}
\usepackage{graphicx}

\title{Paraphrasing Verbs for Noun Compound Interpretation}

\name{Preslav Nakov}

\address{Linguistic Modeling Department, Institute for Parallel Processing\\
         Bulgarian Academy of Sciences\\
         25A, Acad. G. Bonchev St., 1113 Sofia, Bulgaria\\
        {\tt nakov@lml.bas.bg}}

\abstract{
    An important challenge for the automatic analysis
    of English written text is the abundance of noun compounds:
    sequences of nouns acting as a single noun.
    In our view, their semantics is best characterized
    by the set of all possible paraphrasing verbs,
    with associated weights, e.g., \emph{malaria mosquito}
    is \emph{carry (23)}, \emph{spread (16)}, \emph{cause (12)},
    \emph{transmit (9)}, etc.
    Using Amazon's Mechanical Turk, we collect paraphrasing verbs
    for 250 noun-noun compounds previously proposed in the linguistic literature,
    thus creating a valuable resource for noun compound interpretation.
    Using these verbs, we further construct a dataset
    of pairs of sentences representing a special kind of textual entailment task,
    where a binary decision is to be made about whether
    an expression involving a verb and two nouns
    can be transformed into a noun compound, while preserving the sentence meaning.
}

\begin{document}

\maketitleabstract

\section{Introduction}
\label{sec:intro}

An important challenge for the automatic analysis
of English written text is posed by noun compounds --
sequences of nouns acting as a single noun\footnote{This is
\newcite{downing:1977:nc:sem}'s definition of noun compounds.},
e.g., {\it colon cancer tumor suppressor protein} --
which are abundant in English:
\newcite{Baldwin:Tanaka:2004} calculated that noun compounds comprise
3.9\% and 2.6\% of all tokens in the {\it Reuters corpus}
and the {\it British National Corpus}\footnote{There are
256K distinct noun compounds out of the 939K distinct wordforms
in the 100M-word \emph{British National Corpus}.}, respectively.


Understanding noun compounds' syntax and semantics is difficult but important
for many natural language applications (NLP)
including 
question answering, machine translation, information retrieval,
and information extraction.
For example, a question answering system might need to determine whether
`{\it protein acting as a tumor suppressor}' is a good paraphrase
for {\it tumor suppressor protein},
and an information extraction system might need to decide whether
{\it neck vein thrombosis} and {\it neck thrombosis} could possibly co-refer
when used in the same document.
Similarly, a machine translation system facing the unknown noun compound
{\it WTO Geneva headquarters} might benefit from being able to paraphrase it as
{\it Geneva headquarters of the WTO} or as {\it WTO headquarters located in Geneva}.
Given a query like {\it migraine treatment}, an information retrieval system
could use suitable paraphrasing verbs like {\it relieve} and {\it prevent}
for page ranking and query refinement.

\section{Noun Compound Interpretation}
\label{sec:nc}

\begin{table}
  \centering
\begin{tabular}{@{}ll@{ }c@{ }l@{}}
  \multicolumn{1}{c}{\bf RDP} & \multicolumn{1}{c}{\bf Example} &
  \multicolumn{1}{l}{\bf Subj/obj} & \multicolumn{1}{l}{\bf Traditional Name}\\
  \hline
  \texttt{CAUSE$_1$} & {\it tear gas} & object & causative\\
  \texttt{CAUSE$_2$} & {\it drug deaths} & subject & causative\\
  \texttt{HAVE$_1$}  & {\it apple cake} & object & possessive/dative\\
  \texttt{HAVE$_2$}  & {\it lemon peel} & subject & possessive/dative\\
  \texttt{MAKE$_1$}  & {\it silkworm} & object & productive/composit.\\
  \texttt{MAKE$_2$}  & {\it snowball} & subject & productive/composit.\\
  \texttt{USE}       & {\it steam iron} & object & instrumental\\
  \texttt{BE}        & {\it soldier ant} & object & essive/appositional\\
  \texttt{IN}        & {\it field mouse} & object & locative\\
  \texttt{FOR}       & {\it horse doctor} & object & purposive/benefactive\\
  \texttt{FROM}      & {\it olive oil} & object & source/ablative\\
  \texttt{ABOUT}     & {\it price war} & object & topic\\
  \hline
\end{tabular}
  \caption{{\bf Levi's recoverably deletable predicates (RDPs).}
  Column 3 shows modifier's function in the relative clause.}
  \label{table:RDPs}
\end{table}

The dominant view in theoretical linguistics is that noun compound semantics
can be expressed by a small set of abstract relations.
For example, in the theory of \newcite{levi:1978},
complex nominals (a more general notion than noun compounds)
can be derived by two processes -- predicate deletion
(e.g., \emph{pie made of apples} $\rightarrow$ \emph{apple pie})
and predicate nominalization
(e.g., \emph{the President refused general MacArthur's request}
$\rightarrow$ \emph{presidential refusal}).
The former can only delete the 12 abstract recoverably deletable predicates (RDPs)
shown in Table \ref{table:RDPs}.

Similarly, in the theory of \newcite{warren:1978},
noun compounds can express six major types of semantic relations
(which are further divided into finer sub-relations):
\texttt{Constitute}, \texttt{Possession}, \texttt{Location},
\texttt{Purpose}, \texttt{Activity-Actor}, and \texttt{Resemblance}.

A similar view is dominant in computational linguistics.
For example,
\newcite{Nastase:Szpakowicz:2003} use 30 fine-grained relations
(e.g., \texttt{Cause}, \texttt{Effect}, \texttt{Purpose},
\texttt{Frequency}, \texttt{Direction}, \texttt{Location}),
grouped into 5 coarse-grained super-relations:
\texttt{QUALITY}, \texttt{SPATIAL},
\texttt{TEMPORALITY}, \texttt{CAUSALITY}, and \texttt{PARTICIPANT}.
Similarly, \newcite{girju:2005:on:the:semantics} propose a set of 21 abstract relations
(e.g., \texttt{CAUSE}, \texttt{INSTRUMENT}, \texttt{PURPOSE}, \texttt{RESULT}),
and \newcite{Rosario:Hearst:2001:nc:sem} use 18 abstract domain-specific biomedical relations
(e.g., \texttt{Defect}, \texttt{Material}, \texttt{Person Afflicted}).

An alternative view 
is held by \newcite{lauer:1995:thesis},
who defines the problem of noun compound interpretation
as predicting which among the following eight prepositions
best paraphrases the target noun compound:
\emph{of}, \emph{for}, \emph{in}, \emph{at}, \emph{on},
\emph{from}, \emph{with}, and \emph{about}.
For example, \emph{olive oil} is \emph{oil from olives}.

Lauer's approach is attractive since it is simple and
yields prepositions representing paraphrases directly usable
in NLP applications. However, it is also problematic
since mapping between prepositions and abstract relations is hard \cite{girju:2005:on:the:semantics},
e.g., \emph{in}, \emph{on}, and \emph{at}, all can refer
to both \texttt{LOCATION} and \texttt{TIME}.

Using abstract relations like \texttt{CAUSE} is problematic as well.
First, it is unclear which relation inventory is the best one.
Second, being both abstract and limited, such relations capture
only part of the semantics,
e.g., classifying \emph{malaria mosquito} as \texttt{CAUSE}
obscures the fact that mosquitos do not directly cause malaria,
but just transmit it.
Third, in many cases, multiple relations are possible,
e.g., in Levi's theory, \emph{sand dune}
can be interpreted as both \texttt{HAVE} and \texttt{BE}.


Some of these issues are addressed by \newcite{finin:1980:nc:sem:thesis},
who proposes to use a specific verb, e.g.,
\emph{salt water} is interpreted as \emph{dissolved in}.
In a number of publications
\cite{Nakov:Hearst:2006:using:verbs,Nakov:thesis,nakov:hearst:2008:acl},
we introduced and advocated an extension of this idea,
where 
noun compounds are characterized by the set of
all possible paraphrasing verbs, with associated weights,
e.g., {\it malaria mosquito} is
{\it carry (23)}, {\it spread (16)}, {\it cause (12)}, {\it transmit (9)}, {\it etc}.
These verbs are fine-grained, directly usable as paraphrases,
and using multiple of them for a given noun compound approximates its semantics better.

%


Following this line of research,
below we 
present two noun compound interpretation datasets
which use human-derived paraphrasing verbs and
are consistent with the view of an infinite inventory of relations.
By making these resources publicly available,
we hope to inspire further research in paraphrase-based
noun compound interpretation.

\section{Manual Annotations}
\label{sec:mturk}


We used a subset of the 387 examples listed in the appendix of \cite{levi:1978}.
As we mentioned above, Levi's theory targets complex nominals,
which include not only nominal compounds (e.g., {\it peanut butter}, {\it snowball}),
but also nominalizations (e.g., {\it dream analysis}), and
nonpredicate noun phrases (e.g., {\it electric shock}).
We kept the former two categories since they are composed of nouns only and
thus are noun compounds under our definition,
but we removed the nonpredicate noun phrases, which have an adjectival modifier.
We further excluded all concatenations (e.g., {\it silkworm}),
thus ending up with 250 noun-noun compounds.



We then defined a 
paraphrasing task
which asks human subjects to produce verbs, possibly followed by prepositions,
that could be used in a paraphrase involving {\it that}.
For example, {\it come from}, {\it be obtained from}, and {\it be from}
are good paraphrasing verbs for {\it olive oil}
since they can be used in paraphrases like
`{\it oil that \underline{comes from} olives}',
`{\it oil that \underline{is obtained from} olives}'
or `{\it oil that \underline{is from} olives}'.
Note that this task definition implicitly allows for prepositional paraphrases
when the verb is to {\it be} and is followed by a preposition. For example,
the last paraphrase above is equivalent to `{\it oil \underline{from} olives}'.

In an attempt to make the task as clear as possible and to ensure high quality of the results,
we provided detailed instructions, we stated explicit restrictions, and we gave several example paraphrases.
We instructed the participants to propose at least three paraphrasing verbs
per noun-noun compound, if possible.
We used the {\it Amazon Mechanical Turk} Web service\footnote{\texttt{http://www.mturk.com}},
which represents a cheap and easy way to recruit subjects for various tasks that require human intelligence;
it provides an API allowing a computer program to ask a human to perform a task and return the results.

We randomly distributed the noun-noun compounds into groups of 5
and we requested 25 different human subjects per group.
We had to reject some of the submissions,
which were empty or were not following the instructions,
in which cases we requested additional workers in order
to obtain about 25 good submissions per HIT (Human Intelligence
Task).
Each human subject was allowed to work on any number of groups,
but was not permitted to do the same one twice,
which is controlled by the {\it Amazon Mechanical Turk} Web Service.
A total of 174 different human subjects produced 19,018 verbs.
After removing the empty and the bad submissions, and after normalizing the verbs,
we ended up with 17,821 verb annotations for the 250 examples.
See \newcite{Nakov:thesis} for further details
on the process of extraction and cleansing.

\section{Lexicons of Paraphrasing Verbs}

We make freely available three lexicons of paraphrasing verbs
for noun compound interpretation:
two generated by human subjects recruited with {\it Amazon Mechanical Turk},
and a third one automatically extracted from the Web,
as described in \cite{nakov:hearst:2008:acl}.

\subsection{Human-Proposed: All}
\label{sec:mturk:all}

The dataset is provided as a text file
containing a separate line for each of the 250 noun-noun compounds,
ordered lexicographically.
Each line begins with an example number (e.g., 94),
followed by a noun compound (e.g., \emph{flu virus}),
the original Levi's RDP (e.g., \texttt{CAUSE$_1$}; see Table \ref{table:RDPs}),
and a list of paraphrasing verbs.
The verbs are separated by a semicolon and each one is followed in parentheses
by a count indicating the total number of distinct human annotators
that proposed it. Here is an example line:

\vspace{6pt}
\begin{scriptsize}
\noindent \texttt{94	flu	virus	CAUSE1	cause(19); spread(4); give(4); result in(3); create(3); infect with(3); contain(3); be(2); carry(2); induce(1); produce(1); look like(1); make(1); incubate into(1); exacerbate(1); turn into(1); happen from(1); transmit(1); be made of(1); involve(1); generate(1); breed(1); be related to(1); sicken with(1); lead to(1); intensify be(1); disseminate(1); come from(1); be implicated in(1); appear(1); instigate(1); be conceived by(1); bring about(1)}
\end{scriptsize}

\subsection{Human-Proposed: First Only}
\label{sec:mturk:first:only}

As we mentioned above,
the human subjects recruited to work on {\it Amazon Mechanical Turk} (workers)
were instructed to provide at least three paraphrasing
verbs per noun-noun compound.
Sometimes this was hard, and many introduced some bad verbs
in order to fulfill this requirement.
Assuming that the very first verb 
is the most likely one to be correct,
we created a second dataset in the same format,
where only the first verb from each worker is considered.
For example, line 94 in that new text file becomes:

\vspace{6pt}
\begin{scriptsize}
\noindent \texttt{94	flu	virus	CAUSE1	cause(17), be(1), carry(1), involve(1), come from(1)}
\end{scriptsize}

\subsection{Automatically Extracted from the Web}
\label{sec:web}

Finally, we provide a text file in the same format,
where the verbs are automatically extracted from the Web
using the method described in \cite{nakov:hearst:2008:acl}.
This dataset might be found useful by other researchers for comparison purposes.
The corresponding line 94 in that file starts as follows
(here truncated due to a very long tail):

\vspace{6pt}
\begin{scriptsize}
\noindent \texttt{94	flu	virus	CAUSE1	cause(906); produce(21);
give(20); differentiate(17); be(16); have(13); include(11); spread(7); mimic(7);
trigger(6); induce(5); develop from(4); be like(4); be concealed by(3);
be characterized by(3); bring(3); carry(3); become(3); be associated with(3);
$\ldots$}
\end{scriptsize}

\subsection{Comparing the Human-Proposed and the Program-Generated Paraphrasing Verbs}
\label{sec:human}

\begin{figure}
\begin{center}
   \includegraphics[width=230pt]{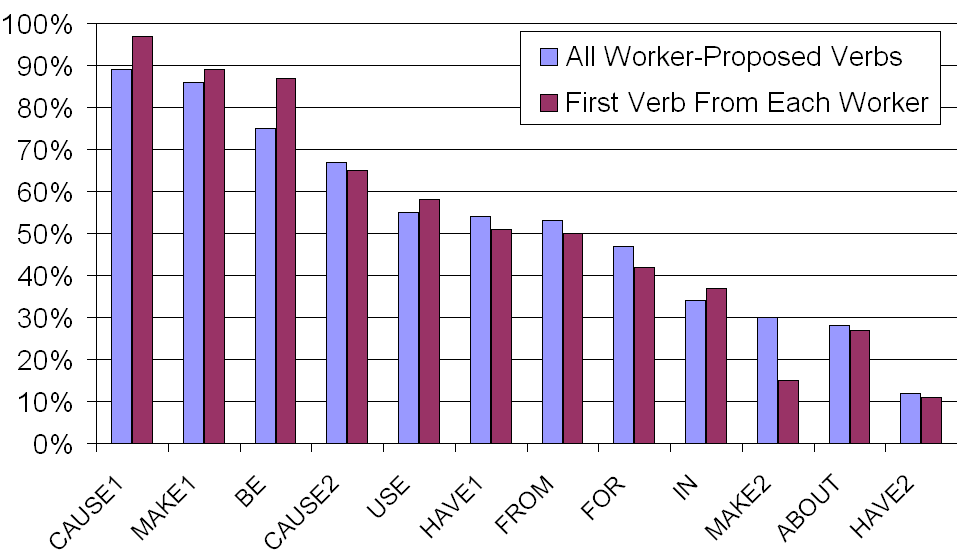}
   \caption{\textbf{Cosine correlation (in \%s) between the human- and the program-
        generated verbs by Levi's RDP:}
        using all human-proposed verbs vs. using the first verb from each worker only.}
\end{center}
   \label{Fig:mturk:by:class}
\end{figure}

In this section, we describe a comparison
of the human- and the program-generated verbs aggregated by Levi's RDP
(see Table \ref{table:RDPs}).
Given an RDP like \texttt{HAVE$_1$}, we collected all verbs belonging
to noun-noun compounds from that RDP together with their frequencies.
From a vector-space model point of view, we summed their corresponding frequency vectors.
We did this separately for the human- and the program-generated verbs,
and we compared them for each RDP.
Figure \ref{Fig:mturk:by:class} shows the cosine correlations
(in \%s) between the human- and the program-generated verbs by Levi's RDP:
using all human-proposed verbs vs. using the first verb from each worker only.
As we can see, there is a very-high correlation (mid 70s to mid 90s) for RDPs
like \texttt{CAUSE$_1$}, \texttt{MAKE$_1$}, and \texttt{BE},
but low correlation 11-30\% for reverse RDPs like \texttt{HAVE$_2$} and \texttt{MAKE$_2$},
and for rare RDPs like \texttt{ABOUT}.
Interestingly, using the first verb only improves the results
for RDPs with high cosine correlation,
but damages low-correlated ones.
This suggests that when the RDP is more homogeneous,
the first verbs proposed by the workers are good enough
and the following ones only introduce noise,
but when it is more heterogeneous,
the additional verbs are more likely to be useful.

We also performed an experiment using the verbs as features
in a nearest-neighbor classifier,
trying to predict the Levi's RDP the noun compound belongs to.
We first filtered out all nominalizations, thus obtaining 214 noun compounds,
each annotated with one of the 12 RDPs shown in Table \ref{table:RDPs},
and we then used this dataset in a leave-one-out cross-validation. 
Using all human-proposed verbs, we achieved 73.71\%$\pm$6.29\% accuracy
(here we also show the confidence interval).
For comparison, using Web-derived verbs and prepositions only yields
50.47\%$\pm$6.68\% accuracy. 
Therefore, we can conclude that the performance with human-proposed verbs
is an upper bound on what can be achieved with Web-derived ones.
See \cite{nakov:hearst:2008:acl} for additional details.

\section{A Dataset for Textual Entailment}
\label{sec:entailment}

Collecting this dataset was motivated by the Pascal Recognizing Textual Entailment (RTE)
Challenge,\footnote{\texttt{www.pascal-network.org/Challenges/RTE2/}}
which addresses a generic semantic inference task
arguably needed by many NLP applications, including question
answering, information retrieval, information extraction,
and multi-document summarization.
Given two textual fragments, a text $T$ and a hypothesis $H$, the goal is
to recognize whether the meaning of $H$ is entailed (can be inferred)
from the meaning of $T$. 
Or, as the RTE2 task definition puts it:

\begin{quote}
``{\it We say that $T$ entails $H$ if, typically, a human
reading $T$ would infer that $H$ is most likely true. This somewhat
informal definition is based on (and assumes) common human
understanding of language as well as common background knowledge.}''
\end{quote}

In many cases, solving such entailment problems requires deciding
whether a noun compound can be paraphrased in a particular way.

The sentences in our Textual Entailment dataset are collected from the Web
and involve some of the above-described human-derived paraphrasing verbs.
These sentences are further manually annotated and provided
in format that is similar to that used by RTE.
Each example consists of three lines, all starting with the example number.
The first line continues with \texttt{\small T:} (the text),
followed by a sentence where the target nouns involved in a paraphrase are marked.
The second line continues with \texttt{\small H:} (the hypothesis),
followed by the same sentence but with the paraphrase re-written as a noun compound.
The third line continues with \texttt{\small A:} (the answer),
followed by either \texttt{\small YES} or \texttt{\small NO},
depending on whether \texttt{\small T} implies \texttt{\small H}.

The following example is positive since
{\it professors that are women} is an acceptable paraphrase
of the noun compound \emph{women professors}:
\vspace{6pt}

\begin{small}
\noindent \texttt{17 T: I have friends that are organizing to get more <e2>\textbf{professors}</e2> \underline{that are} <e1>\textbf{women}</e> and educate women to make specific choices on where to get jobs.}

\noindent \texttt{17 H: I have friends that are organizing to get more <e1>\textbf{women}</e1> <e2>\textbf{professors}</e2> and educate women to make specific choices on where to get jobs.}

\noindent \texttt{17 A: YES}
\vspace{6pt}
\end{small}

The example below however is negative
since a bad paraphrasing verb is used in the first sentence:

\begin{small}
\vspace{6pt}
\noindent \texttt{18 T: As McMillan collected, she also quietly gave, donating millions of dollars to create scholarships and fellowships for black Harvard Medical School students, African filmmakers, and MIT <e2>\textbf{professors}</e2> \underline{who study} <e1>\textbf{women}</e1> in the developing world.}

\noindent \texttt{18 H: As McMillan collected, she also quietly gave, donating millions of dollars to create scholarships and fellowships for black Harvard Medical School students, African filmmakers, and MIT <e1>\textbf{women}</e1> <e2>\textbf{professors}</e2> in the developing world.}

\noindent \texttt{18 A: NO}
\vspace{6pt}
\end{small}

Here is another kind of negative example,
where the semantics is different due to a different phrase attachment.
The first sentence refers to the action of giving,
while the second one refers to the process of transfusion:

\begin{small}
\vspace{6pt}
\noindent \texttt{19 T: Rarely, the disease is transmitted via transfusion of blood products from a <e2>\textbf{donor}</e2> \underline{who gave} <e1>\textbf{blood}</e1> during the viral incubation period.}

\noindent \texttt{19 H: Rarely, the disease is transmitted via transfusion of blood products from a <e1>\textbf{blood}</e1> <e2>\textbf{donor}</e2> during the viral incubation period.}

\noindent \texttt{19 A: NO}
\vspace{6pt}
\end{small}

\section{Conclusion}
\label{sec:conclusion}

We have presented several novel resources 
consistent with the idea of characterizing noun compound semantics
by the set of all possible paraphrasing verbs. 
These verbs are fine-grained, directly usable as
paraphrases, and using multiple of them for a given noun
compound approximates its semantics better.
By making these resources publicly available,
we hope to inspire further research in the direction
of paraphrase-based noun compound interpretation,
which opens the door to practical applications in a number of NLP tasks
including but not limited to machine translation, text summarization,
question answering, information retrieval, textual entailment, relational similarity, etc.

Unfortunately, the present situation with noun compound interpretation is similar
to the situation with word sense disambiguation: in both cases,
there is a general agreement that the research is important and much needed,
there is a growing interest in performing further research,
and a number of competitions are being organized,
e.g., as part of SemEval \cite{Girju:al:2007:semeval}.
Still, despite that research interest, there is a lack of actual NLP applications
using noun compound interpretation, with the notable exceptions
of \newcite{Tatu:Moldovan:2005:entailment} and \newcite{Nakov:2008:WMT},
who demonstrated improvements on textual entailment and machine translation,
respectively. We believe that demonstrating more successful applications
in real NLP problems is key for the advancement of the field,
and we hope that other researchers will find
the resources we release here helpful in this respect.

\section{License}
\label{sec:license}

All datasets are released under the
\emph{Creative Commons License}\footnote{http://creativecommons.org/}.


\bibliographystyle{lrec2006}
\bibliography{bibliography}

\end{document}